%% file: root.tex
\documentclass[letterpaper, 10 pt, conference]{ieeeconf}  

\IEEEoverridecommandlockouts                              

\usepackage{amsmath} 
\usepackage{amssymb}
\usepackage{amsfonts}
\usepackage{amstext}

\DeclareMathOperator*{\argmin}{arg\,min}

\usepackage{mathtools}

\usepackage{graphicx}
\usepackage{xcolor}

\usepackage{booktabs}
\usepackage{multirow}

\usepackage{wrapfig}

\definecolor{Burgundy}{RGB}{144,0,32}

\title{\LARGE \bf
Versatile Demonstration Interface: Toward More \\ Flexible Robot Demonstration Collection
}

\author{Michael Hagenow$^{1}$, Dimosthenis Kontogiorgos$^{1}$, Yanwei Wang$^{1}$, and Julie Shah$^{1}$
\thanks{*This work was supported in part by the MIT Postdoctoral Fellowship Program for Engineering Excellence (PFPFEE) and the Wallenberg Foundation PostDoctoral Research Fellowship.}
\thanks{$^{1}$Michael Hagenow, Dimosthenis Kontogiorgos, Yanwei Wang, and Julie Shah are with the Computer Science and Artificial Intelligence Lab (CSAIL), Massachusetts Institute of Technology, Cambridge, MA 02139, USA.
        {\tt\footnotesize \{hagenow,dimos,felixw,julie\_a\_shah\}@csail.mit.edu}}%
}

\begin{document}

\maketitle
\thispagestyle{empty}
\pagestyle{empty}

\input{00_abstract}
\input{01_introduction}

\input{02_related_work}
\input{03_design}

\input{04_evaluation_v2}
\input{05_discussionv2}

\bibliographystyle{IEEEtran}
\bibliography{references}

\end{document}

%% file: 00_abstract.tex
\begin{abstract}
    Previous methods for Learning from Demonstration leverage several approaches for a human to teach motions to a robot, including teleoperation, kinesthetic teaching, and natural demonstrations. However, little previous work has explored more general interfaces that allow for multiple demonstration types. Given the varied preferences of human demonstrators and task characteristics, a flexible tool that enables multiple demonstration types could be crucial for broader robot skill training. In this work, we propose Versatile Demonstration Interface (VDI), an attachment for collaborative robots that simplifies the collection of three common types of demonstrations. Designed for flexible deployment in industrial settings, our tool requires no additional instrumentation of the environment. Our prototype interface captures human demonstrations through a combination of vision, force sensing, and state tracking (e.g., through the robot proprioception or AprilTag tracking). Through a user study where we deployed our prototype VDI at a local manufacturing innovation center with manufacturing experts, we demonstrated VDI in representative industrial tasks. Interactions from our study highlight the practical value of VDI's varied demonstration types, expose a range of industrial use cases for VDI, and provide insights for future tool design.
\end{abstract}

%% file: 01_introduction.tex
\section{Introduction}
\begin{figure}[t]
\centering
\includegraphics[width=0.98\columnwidth]{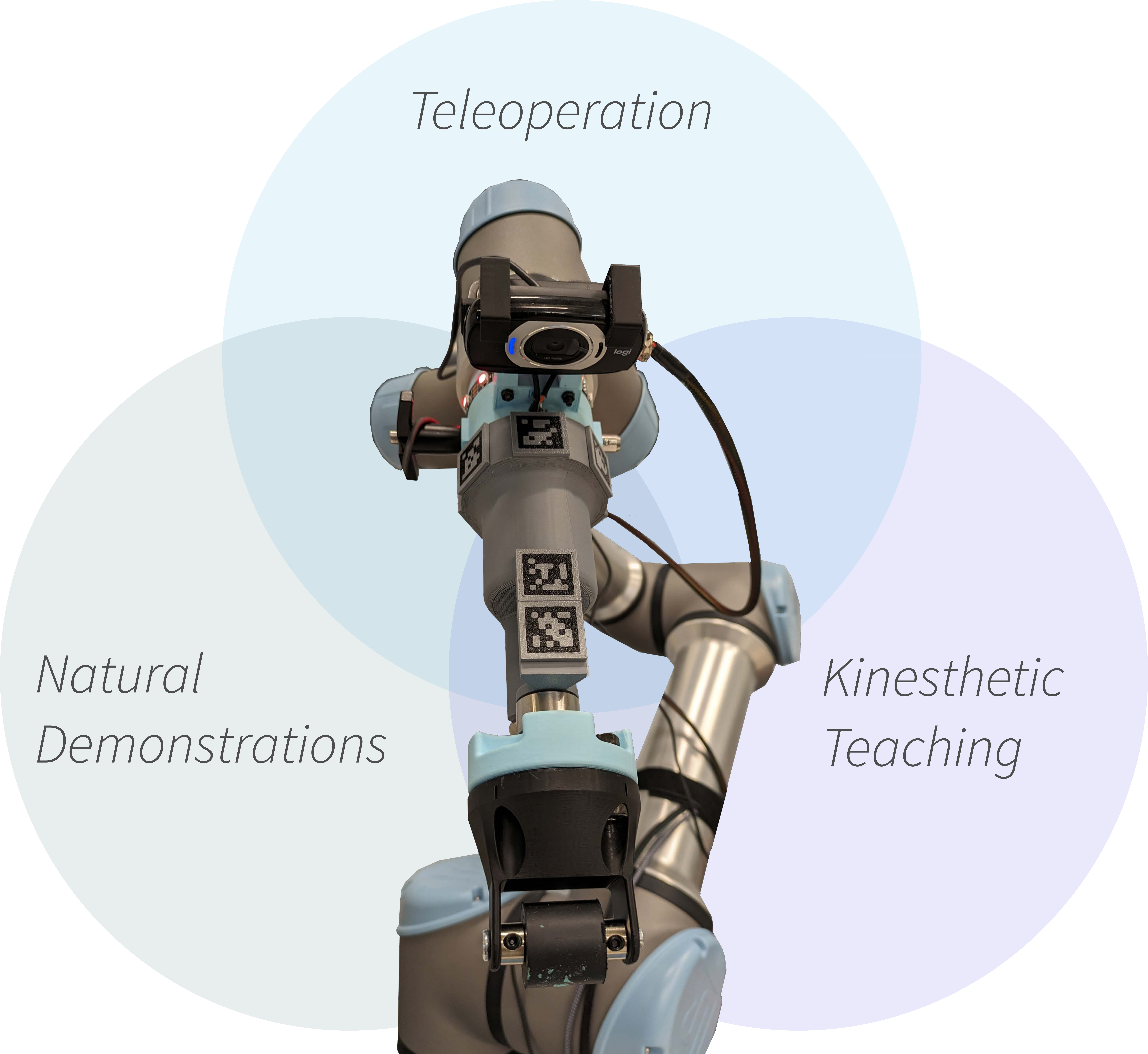}
\vspace{-5pt}
\caption{\textit{Versatile Demonstration Interface} is a tool that connects to the end of a collaborative robot and makes it easier to collect demonstration data through teleoperation, kinesthetic teaching, and natural demonstrations (where the tool detaches from and is tracked by the robot).}
\label{fig:teaser}
\vspace{-15pt}
\end{figure}

When teaching robots new skills, it can be difficult to effectively transfer expert knowledge to a robot's behavior. Learning from Demonstration (LfD) \cite{argall2009survey} has emerged as a prevalent strategy for humans to teach such skills, where a person demonstrates a desired motion rather than relying on traditional low-level robot programming. Within LfD, several approaches have emerged for humans to provide demonstrations; including teleoperation, kinesthetic teaching, and natural interfaces \cite{billard2016learning}. Each of these methods has relative advantages and disadvantages that depend on a number of criteria, including task characteristics, individual demonstrator preferences, and sometimes the algorithm that encodes the robot skill. As an example, teleoperation is a convenient approach to collect demonstration data from a distance, but can suffer in terms of demonstration speed and quality depending on the teleoperation input device. On the other hand, natural demonstrations, such as video recordings of humans or instrumented tools, allow humans to show the robot exactly how they would perform the task. However, natural interfaces often suffer from correspondence issues in mapping human demonstrations to the robot's capabilities. Given the varied benefits of different demonstration modalities, we believe that future flexible automation systems \cite{sanneman2021state} should make it easier to provide different types of demonstrations. In this work, we propose a demonstration interface (shown in Figure \ref{fig:teaser}) that enables experts to provide demonstrations through three common demonstration modalities.

\begin{figure*}[t]
\centering
\includegraphics[width=\textwidth]{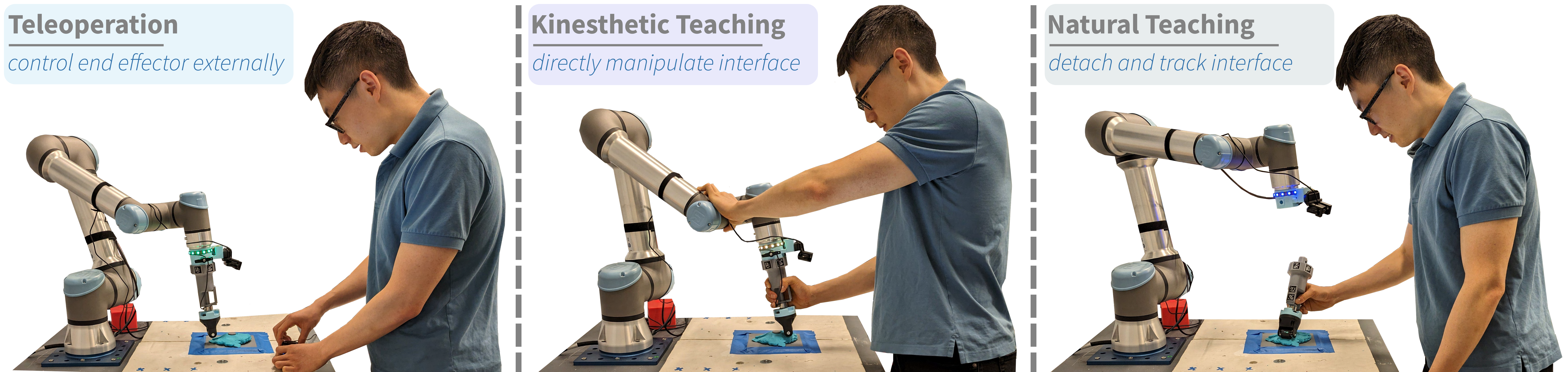}
\vspace{-15pt}
\caption{Examples of the three demonstration modes of MMDI in a rolling task. (Left) The operator remotely controls the robot through a 6D input device. (Middle) The operator directly manipulates the robot arm and the environment forces are measured through a distal force sensor. (Right) The operator detaches the tool and performs demonstrations naturally while the robot optimizes its pose to track the tool with the interface's camera.}
\label{fig:modalities}
\vspace{-15pt}
\end{figure*}

With recent advancements in robot imitation learning, the need for new tools to collect demonstration data is pertinent.
However, there has been little emphasis on creating \emph{general-purpose tools for collecting multiple forms of demonstrations}.  Ideally such general-purpose demonstration tools would be compatible with existing robots, task setups, and provide value to experts in important domains (e.g, manufacturing). Creating a single interface that combines demonstration modalities poses a number of fundamental challenges. This work addresses two key issues: how to design the interface for flexible use between demonstration modalities and how to reduce the effort associated with switching between demonstration types. 

This paper presents \textit{Versatile Demonstration Interface (VDI)}, a robot-mounted interface that enables users to provide three types of demonstrations during LfD. The primary contributions of this paper are the development of a tangible instantiation of VDI and a user study with manufacturing experts that analyzes the potential of VDI's interaction modalities in representative manufacturing tasks. Our results suggest that our implementation of VDI was sufficient to demonstrate representative manufacturing tasks and highlights use cases for VDI's interaction modalities in manufacturing tasks. In the remaining sections, we review related work in the area of LfD, describe the development and features of our interface, and present the findings from a user study where manufacturing experts performed a set of tabletop tasks using the VDI tool.

%% file: 02_related_work.tex
\section{Related Work}
Our flexible LfD interface combines multiple demonstration modalities. To contextualize our contributions, we provide a brief review of LfD methods and previous methods that assess and combine multiple demonstration modalities.

\textbf{Learning from Demonstration Methods --}
From its introduction nearly thirty years ago \cite{bakker1996robot}, Learning from Demonstration (also often called \textit{programming by demonstration}) has gained popularity as a paradigm to teach robots new skills, circumventing many challenges associated with traditional robot programming \cite{argall2009survey}.

Past methods can generally can be categorized into three primary approaches -- kinesthetic teaching, teleoperation, and passive observation (i.e., natural demonstrations) \cite{ravichandar2020recent}. \textit{Kinesthetic teaching}, where a robot arm is guided through training motions, is commercially available in many collaborative robot platforms and been used to teach robots a variety of complex skills \cite{elliott2017learning, schou2013human, kormushev2011imitation, fong2018kinesthetic}. These methods are designed to be intuitive for end users and prioritize feasibility of skill transfer by inherently limiting the demonstrations to motions that can be achieved by the robot. The usability of such methods is often related to the control method, which frequently leverage gravity compensation, compliant, or impedance control \cite{wrede2013user, calinon2009robot}. \textit{Teleoperation}, where a robot is controlled from a distance, is a longstanding robot control paradigm \cite{niemeyer2016telerobotics} that can also be used to collect demonstrations \cite{akgun2012novel}. More recently, vision-based learning systems have leveraged teleoperation to crowd source demonstration data that resembles deployment environments (i.e., with no human occulsions) \cite{padalkar2023open}. \textit{Natural demonstrations}, where a person can perform a demonstration without specific considerations for robot execution, is a logical approach to collect high-quality and realistic human data, but poses many challenges such as correspondence (i.e., how to map human demonstrations to robot capabilities and limits) and state tracking \cite{ravichandar2020recent}. To track the human, many recent methods leverage instrumented tools \cite{akkaladevi2019skill,hagenow2021informing, young2021visual}, augmented reality \cite{soares2021programming,luebbers2021arc}, or learning from observations \cite{stadie2017third,torabi2019recent}.

With recent advances in imitation learning \cite{chi2023diffusion,zhao2023learning}, interest in LfD paradigms for effective data collection have intensified. Recent efforts include new physical interfaces to improve data collection. For example, the Universal Manipulation Interface (UMI) \cite{chi2024universal} is a low-cost natural demonstration interface for collecting prehensile demonstrations. Mobile ALOHA \cite{fu2024mobile}, HATO \cite{lin2024learning}, and GELLO \cite{wu2023gello} are teleoperation setups to control bimanual robot arms with intuitive input devices across diverse household tasks.

\textbf{Modality Comparisons and Fusing Modalities --}
Given the breadth of LfD approaches, previous work has systematically compared demonstration methods. Early comparisons highlighted the value of kinesthetic teaching over more traditional programming approaches (e.g., teach pendants) and teleoperated data collection \cite{akgun2011robot,chernova2014robot,kramberger2014comparison,fischer2016comparison}. More recently, Maric et al. \cite{maric2024comparative} quantified differences in performance and task load when performing contact-rich tasks using kinesthetic teaching and natural demonstrations and Praveena et al. \cite{praveena2019characterizing} compared teleoperation, kinesthetic, hand, and instrumented tool demonstrations across various pick and place tasks. Generally, these previous studies suggest that participants perform better and prefer using more natural input methods (e.g., kinesthetic teaching over traditional programming, hand tracking input over kinesthetic teaching). However, in some cases, task characteristics may drive the choice of demonstration method. For example, a very precise task might benefit from teleoperation, where a user's input can be scaled down (e.g., as is popular in robot-assisted surgery \cite{prasad2004surgical}). Alternatively, when operating a heavy tool, it may be beneficial for the operator to provide kinesthetic guidance over natural demonstrations to avoid the physical burden of the tool (e.g., through robot force offloading) \cite{ajoudani2018progress}. When considering a flexible tool for use in many tasks, we thus see merit in interfaces combining modalities.

While limited work has investigated tools fusing together demonstration modalities, initial results suggest there may be benefits for both user interactions and policy learning. Open Multi-Purpose Interface (OMI) \cite{raei2024multipurpose} combines kinesthetic teaching and teleoperation and exemplifies the benefits of a flexible interface in a multi-step home-care task using a mobile collaborative robot. In policy learning, MimicPlay \cite{wang2023mimicplay} highlights the benefits of combining natural demonstrations with a small amount of teleoperated demonstrations to overcome robot correspondence issues. Similarly, PoCo \cite{wang2024poco} has shown how information from different demonstration modalities can be combined to learn more general skills. While heterogeneous learning is not a focus of this work (we instead focus on human perceptions and use cases for multiple forms of demonstration), we are optimistic that advances in heterogeneous learning will open further opportunities for more general-purpose tools for human demonstrations.

%% file: 03_design.tex
\section{Versatile Demonstration Interface}

To illustrate the opportunities associated with combining multiple demonstration modalities and to collect design and use case feedback, we developed a hardware implementation of VDI. Here we discuss the important features of our physical and interaction design. Example interactions using our prototype interface are shown in the supplemental video.

\subsection{Design and Sensing}
\begin{figure}[]
\centering
\includegraphics[width=\columnwidth]{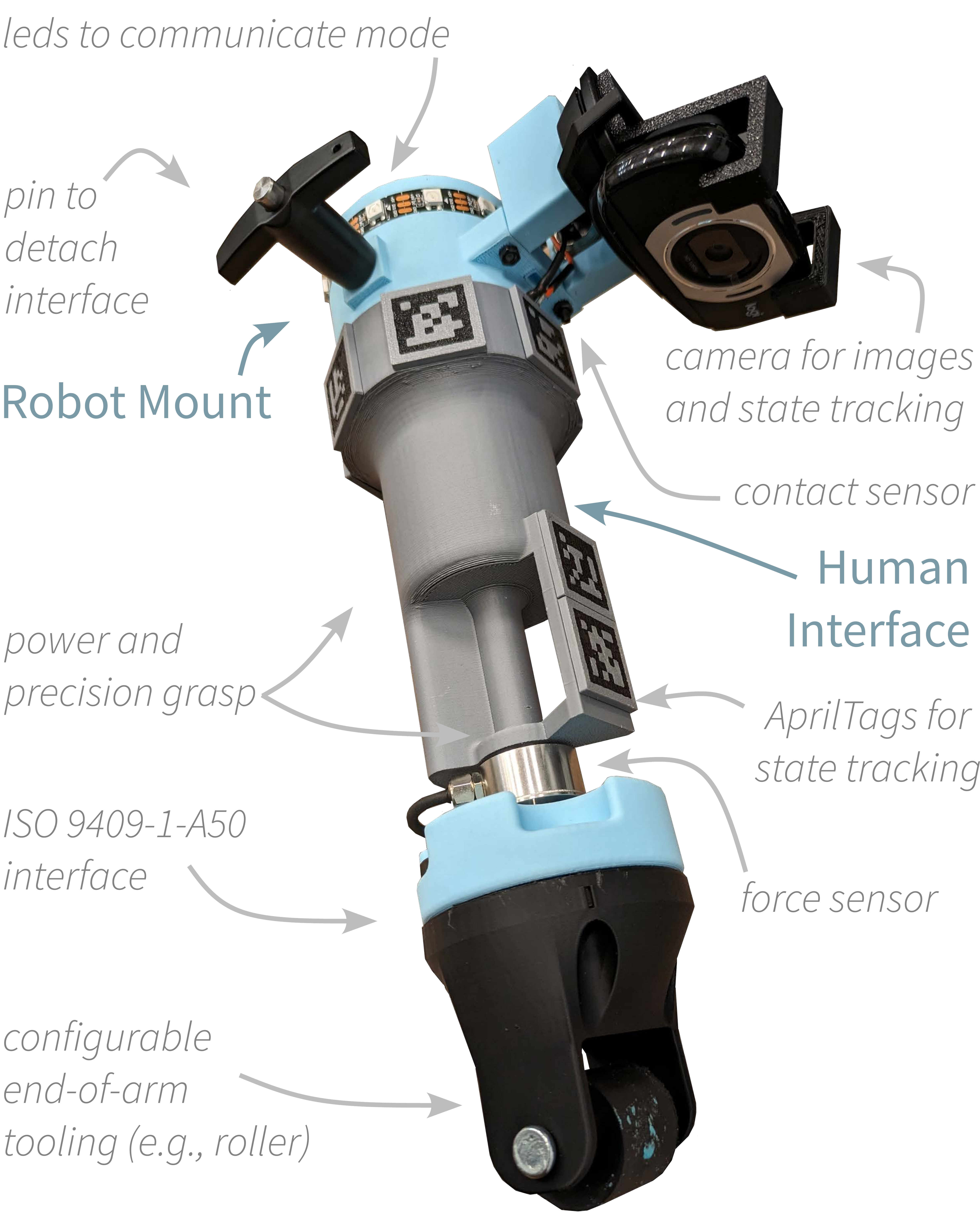}
\vspace{-15pt}
\caption{Schematic of the prototype interface. The interface consists of two main sections: the robot mount and the human interface. The schematic also illustrates the key features supporting the interaction modes.}
\label{fig:design}
\vspace{-15pt}
\end{figure}

Our primary design goals were to (1) enable the three types of common demonstrations, (2) interface with existing end-of-arm tooling, and (3) require only sensing that is typically available on commercial robots or connected to the tool. In our prototype design, we also optimized for human factors when possible by following hand-tool design best practices. Our interface design schematic is shown in Figure \ref{fig:design}.

Our interface was designed as an extension that can be mounted to existing collaborative robot platforms. In our implementation, we mounted our interface to a Universal Robots UR5e robot. To enable flexible use with the variety of existing robot tooling, our interface attaches to the robot with the common DIN ISO 9409-1-A50 and terminates with the same mounting pattern. To minimize the impact to robot kinematics, our interface was designed with a minimal length to accommodate human interaction. In two of the demonstration modalities, the human operator directly grasps the interface. Depending on the task force and precision requirements, it may be desirable to perform the task using either a \emph{power} or \emph{precision} grasp. As shown in Figure \ref{fig:design}, the interface accommodated both types of grasps. The interface geometry followed best practices for hand tool design \cite{radwin1996ergonomics}, such as the grasp diameters ($50\;\texttt{mm}$ for power and $13 \;\texttt{mm}$ for precision) and a flange to offload the tool weight. Future studies of the interface will assess human factors, particularly for varied end-of-arm tooling. We also included an LED ring on the robot mount to communicate with the user (e.g., the current mode and the level of applied force in teleoperation).

To aid in LfD, the required sensing for data collection is included on-board the VDI interface. The interface consists of two main parts: the \emph{robot mount} and the \emph{human interface}. The robot mount includes a camera that can be used for visuomotor policy learning (i.e., collecting images of the tool during the demonstration) and for tracking the interface (e.g., the pose of the tool) when it is detached from the robot. The camera is placed in a fixed location with a known transform for localization to the robot base. For many tasks in industrial manufacturing, such as surface finishing, it is necessary for robots to regulate force in addition to kinematic goals. Thus, the tool includes a force sensor between the human interface and the end effector. This force sensor can be used to track applied forces when the human is grasping the interface. In our implementation, we use a uni-axial load cell sensor as a proof-of-concept force sensor (a 20 kg DYMH-103 Wheatstone bridge sensor with a HX711 amplifier), though such low-cost force sensors can suffer from nonlinearity and crosstalk effects \cite{cao2021six}. We found that the resolution and repeatability was sufficient for our study, where the force information was primarily needed to identify mode switches. Future versions will employ a multi-axis sensor for higher fidelity data collection. A key design element is the ability to remove part of the interface from the robot to directly use it as a hand tool. To accommodate this feature, the robot mount includes a removable pin and force-sensitive resistor (FSR) that is used as a contact sensor. The human interface also employs a pattern of AprilTag \cite{olson2011apriltag} markers with known configurations that are tracked by the robot interface camera.

\subsection{Demonstration Modalities}

\textbf{Teleoperation --} In our implementation, the 6D teleoperation input to the robot is provided through a SpaceMouse, though our approach is amenable to any remote input device. Because the SpaceMouse does not provide force feedback to the operator, we operate our robot in an admittance control mode and use LEDs and sound to communicate when the level of applied force is approaching the set force limit. The gains (i.e., scaling) of the teleoperation input generally trade off precision with speed. In our implementation, we empirically tuned the gains for precise manipulations (as was required by the study tasks). The operator can enter teleoperation mode at any point when the tool is attached by pressing on the remote input device. If the operator exceeds the set force limit, the robot exits the teleoperation mode.

\textbf{Kinesthetic Teaching --} Our implementation leveraged the gravity compensation mode of the UR5e (i.e., free-drive mode) for kinesthetic interactions. We chose free-drive mode as a safe, passive solution for our user study, though depending on the collaborative robot platform, the kinesthetic teaching mode may also be achieved through impedance or admittance control. The tool-side force sensor is used to record interactions with the environment. The discrepancy between the robot force estimate and tool-side force estimate is used as a heuristic for whether to activate the kinesthetic mode. Intuitively, if the operator grabs onto and pulls on the interface, the robot will switch to the kinesthetic mode. In our implementation, the operator was required to pull in the tool direction because of the uni-axial force sensor. Depending on the force required for the task (and to kinesthetically guide the robot), the operator can choose to use either a \textit{power} or \textit{precision grasp} of the interface.

\textbf{Natural Demonstrations --} To engage in the natural demonstration mode, the operator can pull a pin on the robot mount and remove the human interface. Once the contact sensor detects that the tool is removed, the LEDs flash blue to indicate the system is ready to begin the natural mode. As soon as the camera locates one of the interface AprilTags, it will be begin to film the natural demonstration. To track the tool using the robot-mounted camera, we employ a nonlinear optimization of the robot pose that balances viewpoint heuristics, as is common in previous methods for shared control of robot cameras \cite{rakita2018autonomous,senft2022method,li2023stargazer,praveena2023periscope, dreher2022erfassung}.  The optimization is a nonlinear weighted-sum of objectives subject to inequality constraints.

\begin{equation}
\begin{aligned}
    \argmin_{\textbf{p}_r,\textbf{R}_r} \quad & \sum\limits_{i}w_i \phi_i(\textbf{p}_r,\textbf{R}_r,\textbf{p}_h,\textbf{R}_h)\\
    \textrm{s.t.} \quad & g_i(\textbf{p},\boldsymbol{\omega})\leq 0 \;\forall i \in \{1,..,m\}
\end{aligned}
\end{equation}
where $\textbf{p}_r$ is the commanded position of the robot camera, $\textbf{R}_r$ is the robot camera rotation matrix, $\textbf{p}_h$ is the estimated position of the human interface, $\textbf{R}_h$ is the estimated rotation matrix of the human interface, $w_i$ are the objective weights, $\phi_i$ are the objectives, and $g_i$ are the $m$ inequality constraints. While many viewpoint heuristics are possible, we leveraged four objectives that yielded safe (for the user study) and appropriate behaviors. We provide the objectives and parameters for reproducibility, but acknowledge that different robots and workspaces may require different objectives or parameters. Our robot camera pose objectives optimize:

\begin{itemize}[leftmargin=*]
\item  Maintaining an appropriate distance between the robot camera and the human interface.
\begin{equation}
    \phi_1(\textbf{p}_r,\textbf{R}_r,\textbf{p}_h,\textbf{R}_h) = \left( (\textbf{R}_r(\textbf{p}_h -\textbf{p}_r))^\textrm{T}\hat{\textbf{z}}-d\right)^2
\end{equation}
where $\hat{\textbf{z}}=[0,0, 1]^\textrm{T}$ is a unit vector that extracts the distance relative to the camera z-axis and $d$ is the desired camera-interface distance. In our implementation, $w_1=100$ and $d=0.3\textrm{m}$. The distance was selected to balance field-of-view and AprilTag recognition. 
\item Centering the tool in the camera view by aligning the camera with the vector between the camera and interface.
\begin{equation}
\phi_2(\textbf{p}_r,\textbf{R}_r,\textbf{p}_h,\textbf{R}_h) = \left((\frac{\textbf{p}_h -\textbf{p}_r}{||\textbf{p}_h -\textbf{p}_r ||})^\textrm{T}(\textbf{R}_r\hat{\textbf{z}})-1\right)^2
\end{equation}
In this objective, $w_2\!=\!100$. We also considered an additional objective that constrained the viewing angle of the centered tool (for better viewing of the AprilTags), but found it had little impact on behavior empirically.
\item Encouraging a neutral (i.e., centered) camera pose and angle that we found through pilots helps to avoid local minima in the optimization.
\begin{gather}
    \phi_3(\textbf{p}_r,\textbf{R}_r,\textbf{p}_h,\textbf{R}_h) = ||\textbf{p}_r-\textbf{p}_n||^2 \\
    \phi_4(\textbf{p}_r,\textbf{R}_r,\textbf{p}_h,\textbf{R}_h) = \angle(\textbf{R}_n^\textrm{T}\textbf{R}_r)^2
\end{gather}
where $\textbf{p}_n = [0, -0.4, 0.35]^\textrm{T}$ is the neutral camera position, $\textbf{R}_n$ is the neutral orientation, and $\angle(\cdot)$ the magnitude of rotation. In our experimental setup, we only found it empirically beneficial to penalize rotation about the $y$-axis, which corresponded to the principal axis of the robot's workspace (i.e., table). We encouraged the neutral camera pose through small penalties ($w_3\!=\!2$ and $w_4\!=\!0.5$).
\end{itemize}

The inequality constraints (i.e., $g_i$) are used to limit the rotation and commanded position of the robot to within the robot's dexterous workspace. For position, we enforced the following constraint: $[-0.3, -0.45, -0.2] \leq \textbf{p}_h \leq [0.3, -0.25, 0.55]$. For rotation, we constrained rotation about the first two axes: $-0.45\leq\theta_x\leq0.0$ and $-0.8\leq\theta_y\leq0.8$. The $z$-axis rotation was a fixed value such that the camera would face toward the human.
As part of the optimization, we also limited the linear ($\leq1\; \textrm{cm/s}$) and angular velocity ($\leq 0.1\; \textrm{rad/s}$) of the commanded robot pose for participant safety. 
 The state of the tool (i.e., $\textbf{p}_h$ and $\textbf{R}_h$) is estimated using the AprilTags. Each time a camera sees an AprilTag, the known configuration provides an estimate of the 6D tool pose.  The various estimates of the tool pose are used as input to an Extended Kalman Filter (EKF) which provides the final pose estimation. In our implementation, the pose estimation was throttled to 5 Hz to allow for more observations per output pose estimate. For the purposes of our initial prototype implementation and optimizing the camera pose, we found this rate to be sufficient. However, for demonstration data collection, a higher estimation rate will be necessary. We imagine achieving a higher frequency through existing sensor fusion methods (e.g., adding an IMU to the interface). As an alert, the system beeps when no markers are visible.


%% file: 04_evaluation_v2.tex
\begin{figure}[b]
\centering
\includegraphics[width=1.0\columnwidth]{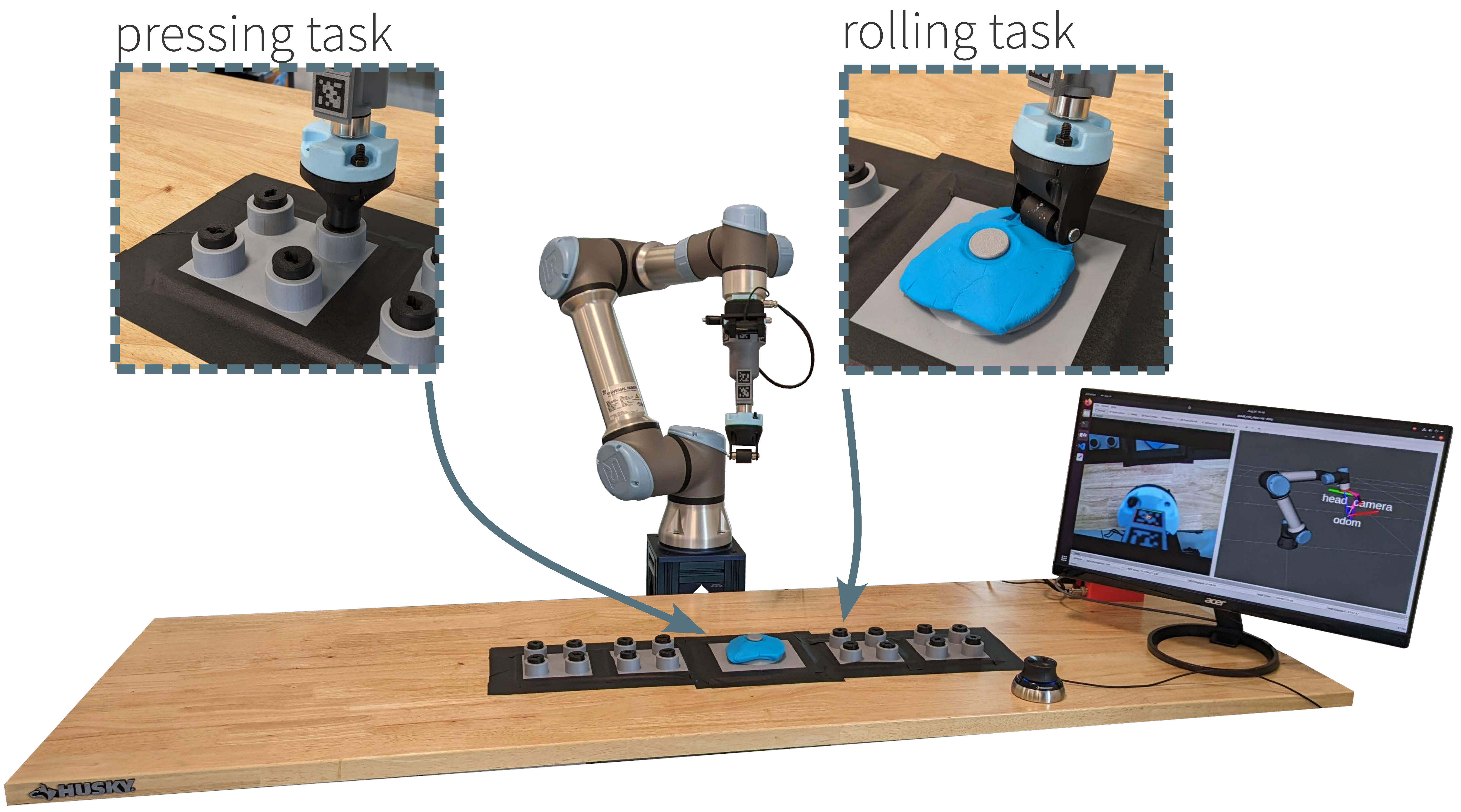}
\vspace{-15pt}
\caption{User evaluation tasks. \textit{Left:} The press-fitting task where participants press sixteen fittings into designated slots. The two groups of fittings were on opposite ends of the workbench.
 \textit{Right:} Rolling task where participants used the end effector to try and evenly distribute the material over the contoured surface. A pin helps to keep the material in place.}
\label{fig:exptasks}
\end{figure}

\begin{figure*}[t]
\centering
\includegraphics[width=\textwidth]{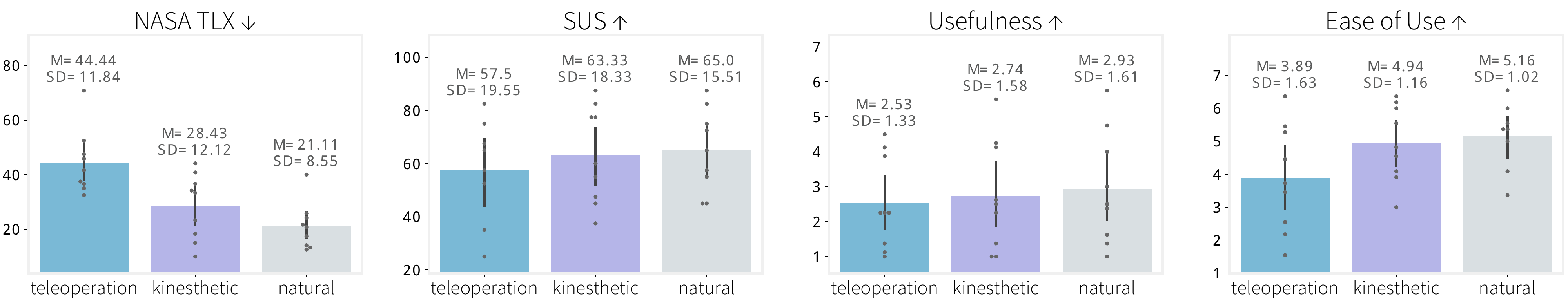}
\vspace{-15pt}
\caption{User study questionnaire results comparing the three demonstration modalities in terms of task load, usability, usefulness, and ease of use.}
\label{fig:study}
\vspace{-15pt}
\end{figure*}

\section{Evaluation}
Our vision is to enable flexible Learning from Demonstration interfaces that can adapt to different demonstration needs (by enabling multiple demonstration modalities) and do not require sensing beyond what is available onboard the robot and LfD interface. To assess our prototype robot interface, we conducted a user study with manufacturing experts. The goals of our study were to (1) assess expert interactions with the robot-tracking based natural demonstration mode to understand its utility and limitations, (2) to probe participants regarding manufacturing applications where each demonstration mode would be valuable (i.e., demonstrating the practical value of a single flexible interface for multiple demonstrations in manufacturing) and (3) to further understand opportunities for manufacturing tasks that would benefit from VDI's ability to easily switch between demonstration modes within a task.

\subsection{Task Description}
We designed two tasks to resemble processes in industrial manufacturing. Both task setups are shown in Figure \ref{fig:exptasks}. The first task was a \textit{rolling} task, which proxies a manual thermomolding task and consisted of rolling a block of play-doh against a contoured mold. The play-doh was initially formed into a puck-like shape and secured to the mold using a pin (for one-handed completion). The purpose of the first task was to expose participants to using the three demonstration modes of VDI in a complex, contact-rich task.

The second task was a \textit{press-fitting} task where patterns of cylindrical fittings were pressed into a mold. The task was designed to combine coarse, fine, and force-controlled motions. Participants pressed sixteen fittings into a mold. The fittings were arranged in groups of eight with the two groups on opposite ends of the workbench. The purpose of the second study task was for participants to explore switching between demonstration modalities and to use VDI under varied task characteristics (e.g., coarser, discrete motions).

\subsection{Manufacturing Study Population}
Our study involved nine experts, summarized in Table \ref{tab:manufacturing-participants}, (7M, 2F), aged 33--69 ($M = 45.3, SD = 11.1$), recruited from the Berkshire Innovation Center via email. Eight participants were from a formal manufacturing background with an average of 15 years of experience. The procedure was administered under a protocol \#2402001226 approved by the Institutional Review Board (IRB) of the Massachusetts Institute of Technology. The experiment was designed to last an hour and a small participation remuneration was paid.

\begin{table}[t]
    \centering
    \small
    \begin{tabular}{@{}p{0.3cm}p{0.3cm}p{0.7cm}p{3.5cm}p{2cm}@{}}
        \toprule
        \textbf{ID} & \textbf{Age} & \textbf{Yrs} & \textbf{Role} & \textbf{Industry} \\
        \midrule
        P1 & 33 & 5 & Maintenance Engineer & Automotive \\
        P2 & 57 & 30 & Industrial Artist & Sculptures \\
        P3 & 34 & 3 & Finance and Management & Food Tech. \\
        P4 & 39 & 20 & Manager & General Mfg \\
        P5 & 42 & 24 & Support Manager & Paper \\
        P6 & 52 & 16 & Ortho. Massage Therapist & Healthcare \\
        P7 & 42 & 18 & R\&D & Plastics \\
        P8 & 40 & 1.5 & Consultant & General Mfg \\
        P9 & 69 & 20 & Plant Manager & Chemicals \\
        \bottomrule
    \end{tabular}
    \caption{Summary of Participants}
    \vspace{-25pt}
    \label{tab:manufacturing-participants}
\end{table}

\textbf{Procedure --} Following informed consent, participants were briefed on the structure of the experiment. The participants then completed the rolling task using each of the conditions. The order of the conditions followed a Latin-square design to balance between controlling for order efforts and replicating order sequences. For each condition, the participant was briefed on how to use the method and given five minutes to interact with the method. After each condition, the participants filled out questionnaires.

After completing the three conditions of the rolling task, the participants completed the press-fitting task. Participants were instructed about the goals of the task and shown how to complete the task without the robot. Participants were prompted that they could switch demonstration modes at any point during the task, but that they needed to use each mode to complete at least one press fitting. Following the two tasks, participants engaged in a semi-structured interview where they were asked about their preferences, design feedback, and potential use cases for our VDI implementation.

\textbf{Metrics --} Most critical to our evaluation objective was the final semi-structured interview.  Responses from the semi-structured interview were synthesized into the key themes presented in the next section. We additionally collected task load, usability, and preference data to confirm that our combined demonstration interface is consistent with modality findings from previous studies \cite{praveena2019characterizing,maric2024comparative}. After each rolling task, participants filled out the NASA Task Load Index (TLX) \cite{hart1988development}, the Systems Usability Scale (SUS) Questionnaire \cite{brooke1996sus}, and the Usefulness and Ease of Use 7-point subscales of the USE Questionnaire \cite{lund2001measuring}. Given our small sample size, we report only descriptive statistics for the questionnaire responses.

%% file: 05_discussionv2.tex
\section{Results and General Discussion}
Here we discuss the results of our user study and takeaways, limitations and future work, and general conclusions. All participants were generally able to complete the rolling task with each demonstration modality. In the press-fitting task, all participants were able to complete the task using the VDI tool. The questionnaire response results are shown in Figure \ref{fig:study} and participants ranked preferences are shown in Table \ref{tab:ranked_preferences}. Two participants responses are omitted from the questionnaire results due to missed survey items.

\subsection{User Study  Interactions and Implications}

\begin{table}[b]
\centering
\begin{tabular}{p{1.6cm}p{1cm}p{1cm}p{1cm}p{1.6cm}}
\hline
\textbf{Method} & \textbf{1st} & \textbf{2nd} & \textbf{3rd} & \textbf{Mean Rank} \\
\hline
Teleoperation & 0 & 4 & 5 & 2.56 \\
Kinesthetic   & 3 & 4 & 2 & 1.89 \\
Natural       & 6 & 1 & 2 & 1.56 \\
\hline
\end{tabular}
\caption{Ranked preferences for demonstration methods}
\vspace{-25pt}
\label{tab:ranked_preferences}
\end{table}

\textbf{The natural demonstration form factor is appropriate, but robot state tracking needs improvements for industrial tasks.} Consistent with previous studies, the majority of participants ranked the natural mode of VDI as their top preference and most favorable in terms of usability, usefulness, and ease of use. Regarding the human interface, we found that all participants chose to use the precision grasp. We were encouraged that several participants commented that the form factor felt familiar, similar to using a pencil [P3] or a manual tool [P1]. Despite the length of the tool (to accommodate force sensing), participants commented that the natural method was still the fastest way to accomplish the tasks [P7], would be preferred for tasks with delicate motions [P4], and that they generally felt they could perform the task better compared to the other demonstration modes [P5, P8]. Participants identified a range of manufacturing applications for natural demonstrations, including molding [P3], industrial food preparation [P3], coordinate metrology [P4], prepping and crimping wires [P5], powder coating [P7], material removal [P9], and sculpting [P9].

\begin{figure}[]
\centering
\includegraphics[width=1.0\columnwidth]{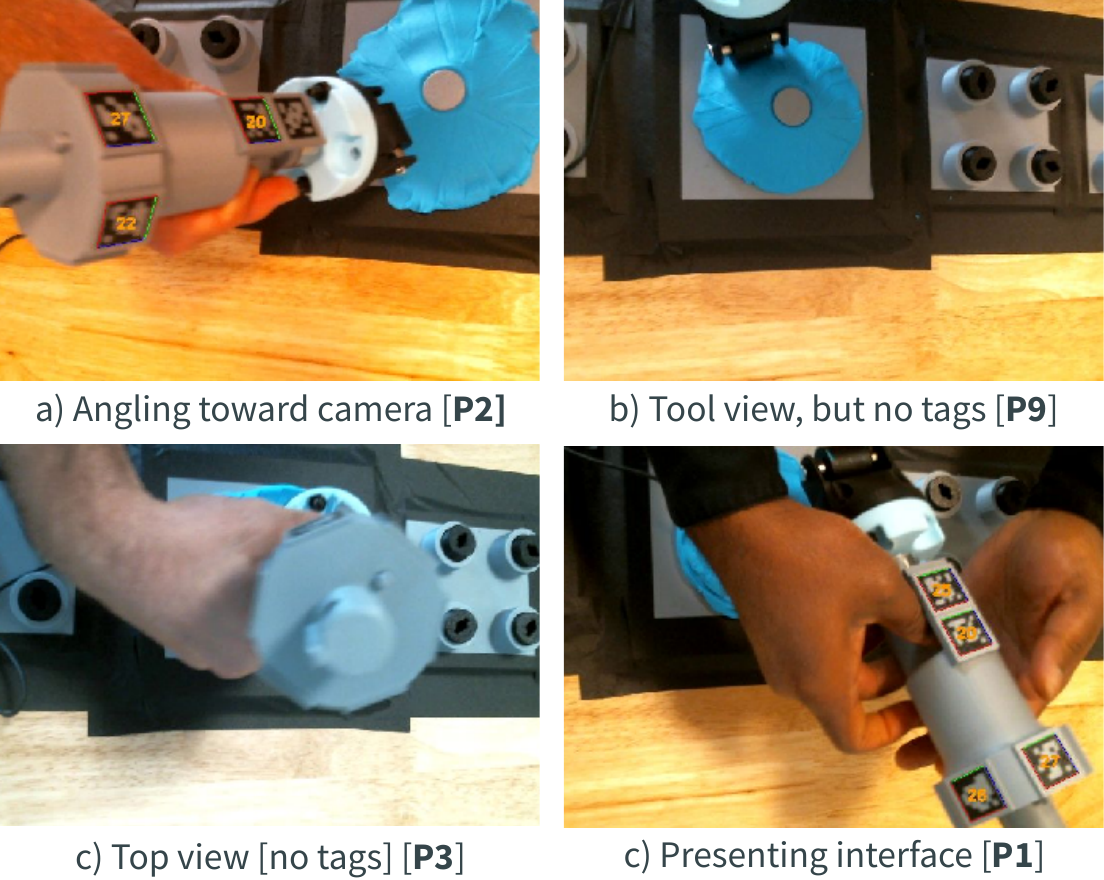}
\vspace{-20pt}
\caption{Examples of successful strategies and failure cases of the natural demonstration mode. a) Several participants angled the front of the tool (with the most AprilTags) toward the robot during the task, which minimized movement of the robot camera and generally improved tool tracking. b) The robot could sometimes see the end effector, but none of the Apriltags which caused tracking failures. c) Occasionally, participants faced the top of the human interface toward the camera which did not include tracking tags. d) Some participants would stop the task and present the tool to the robot when it beeped. While this would temporarily restore state estimation, it led to later tracking issues when participants reoriented the interface to continue.}
\label{fig:qualnatural}
\vspace{-15pt}
\end{figure}

Participants had varied levels of success interacting with the natural mode's camera tracking system. Figure \ref{fig:qualnatural} illustrates common successful participant strategies and failure cases. Many participants found success by avoiding large coarse motions and angling and rotating the tool such that the AprilTags were clearly visible at all times. Empirically, we found that this participant strategy tended to avoid local minima in the camera optimization. Less successful participants tended to exclusively focus on the task and at times ignored the camera optimization beeping indicator. Our optimization contained constraints to limit camera motion to a small envelope for participant safety. In the prototype, participants commented that the \textit{range was very limited} [P6] and that it could at times be difficult to guide the robot to certain views of the task (i.e., the optimization encountered local minima). Given such limitations in range, P3 mentioned the natural mode may be most effective for \textit{2D} (e.g., tabletop) tasks where a similar camera angle can be used to track all hand tool poses. 

From participant interviews and observations during the study, we believe a combination of future design modifications and participant training can further improve VDI's natural demonstration experience. A few participants desired further flexibility to grasp the interface, such as through additional handles [P7] or modifications to accommodate two-handed grasps for better control [P9]. It would also be beneficial to add state-tracking markers on top of the human interface for scenarios where the camera is above the tool, though we assert the camera-pose optimization should try to escape such configurations as they occlude the view of the tool and propagate pose estimation error. Regarding the camera motion constraints, we believe that practical setups can consider wider ranges of motion as long as velocity is limited to minimize unsafe robot impacts with workers. Additionally, P3 suggested using a tracking camera with a wider field of view, which could aid in better state tracking with less robot motion. The main empirical limitation we encountered were situations where the camera would become stuck trying to follow the user's motion. In these cases, participants became frustrated trying to guide the optimization out of the local minima and back to a more favorable pose. We believe that in future iterations, a critical functionality for the natural mode will be the option for the human to manually reposition the robot to a new desired view. Finally, given that several participants arrived at strategies (e.g., tilting the tool) that led to more favorable tracking, we believe that future user education and training can also help improve the overall experience.

\textbf{There are opportunities for each demonstration mode in manufacturing depending on user or task characteristics.} Consistent with previous studies, we found that kinesthetic teaching was generally preferred over teleoperation. We also found that the usability, usefulness, and ease of use was worst for teleoperation and second best for kinesthetic teaching. Despite these general trends, participants thought that operator background might heavily influence desired demonstration modes for manufacturing floor workers. As P7 commented, \textit{``With the engineers that I work with… we have a broad range of right-out-of-school to thirty years’ experience. People would pick different methods for sure."} For example, several participants noted that teleoperation would likely be popular with young manufacturing workers, who might have more video game experience. With our experienced study population, teleoperation was not preferred.

While in some cases, operator differences may govern the choice of demonstration modality, task characteristics would be the more likely driver in manufacturing. As indicated above, the natural mode was preferred for tasks with precision requirements, localized motions (for easier robot tracking), and where it is feasible for the worker to be collocated. For kinesthetic teaching, participants imagined use in manufacturing applications where more direct control was desired, but where the robot could assist with pushing/pulling [P5], or maintaining orientation [P1,P2]. Such assistive applications would make it so \textit{`I didn't have to do too much, but I had some control} [P8]". Participants thought kinesthetic demonstrations would be well-suited for tasks such as screw fittings on a small assembly line [P1], stoppering [P1], tapping holes [P2], grinding [P2], and drilling [P4]. Regarding kinesthetic interactions with VDI, we observed that participants often occluded the robot-mounted camera with their hand or arm. Thus, we believe a future implementation needs either (1) further optimization of the camera placement (e.g., further out from the robot) or (2) indications on the interface for how the interface can be grasped without occlusion.

For teleoperation, participants saw value for tasks where the demonstrator couldn't be safely near the robot, such as environments with bio-hazards [P6] or heavy payloads [P7]. Participants also commented on the value of using teleoperation in manufacturing to scale down human input for small, precise motions [P4, P9] or to avoid the fatigue associated with more manual demonstration methods of repetitive tasks [P4]. Applications such as soldering or assembly with scaled-down motion [P3], corrosive material handling [P4], process monitoring [P7], and high-strength press fits [P9] were identified as good manufacturing applications for teleoperation.

\textbf{There was limited support for manufacturing applications where switching demonstration modes would be valuable within a task.} Several participants mentioned using one demonstration mode for more coarse motions (e.g., kinesthetic teaching) and one mode for more precise motions (e.g., natural or teleoperation) [P2, P3, P9]. However, most participants could not point to a manufacturing application where this behavior would be useful. Thus, we believe the primary value of VDI in manufacturing lies in the interface's flexibility for different users or task characteristics.


\subsection{Study Limitations}
Our study only focused on perception and manufacturing use cases for the interaction modalities of the VDI. While we believe this initial study was critical to understand viability for VDI in manufacturing applications, further iteration on the physical design (following our study takeaways and design guidelines) and testing of VDI is needed. Future studies should focus on improving and characterizing the quality of VDI data collection, investigating impacts of the demonstration modalities (as well as combinations of demonstration data) on robot policy training, and exploring opportunities for robots to elicit targeted demonstration types from operators. Future studies should also test with several robots, more teleoperation input devices, through a wider suite of representative manufacturing tasks, and in domains outside of manufacturing (e.g., medical applications). 

\subsection{Conclusion and Implications for LfD}
We presented \textit{Versatile Demonstration Interface (VDI)}, an interface that attaches to existing collaborative robot platforms and enables operators to teach robots through three types of demonstrations: teleoperation, kinesthetic, and natural. From our study with manufacturing experts, we highlighted the value of natural demonstrations with the robot-tracked instrumented tool, confirmed the utility of VDI in manufacturing settings through varied use cases, and assessed the role of VDI mode switching in manufacturing tasks. Future work will build more mature interfaces for data collection and assess VDI in robot policy learning.